%% file: RigidAnchor-Manuscript-to-Arxiv.tex
\documentclass{article}

\usepackage[preprint, nonatbib]{neurips_2023}

\usepackage[utf8]{inputenc} 
\usepackage[T1]{fontenc}    
\usepackage{url}            
\usepackage{booktabs}       
\usepackage{amsfonts}       
\usepackage{nicefrac}       
\usepackage{microtype}      
\usepackage{xcolor}         
\usepackage{graphicx}
\usepackage{amsmath}
\usepackage{multirow}
\usepackage{marvosym}

\definecolor{linkcolor}{RGB}{255,0,0}
\definecolor{urlcolor}{RGB}{255,105,180}
\definecolor{citecolor}{RGB}{66,168,235}
\usepackage[pagebackref=true,breaklinks=true,colorlinks,bookmarks=false]{hyperref}
\hypersetup{colorlinks=true,linkcolor=linkcolor,urlcolor=urlcolor,citecolor=citecolor}

\input{body/config}

\input{body/head_arxiv}

\begin{document}

\maketitle

\input{body/abstract}

\input{body/body}

\bibliographystyle{ieee_fullname}
\input{body/tail}

\end{document}

%% file: body/config.tex
\graphicspath{{D:/0-paper/20261012论文/素材/图片/}{fig/source/}}

%% file: body/head_arxiv.tex
\newlength{\wcol}
\AtBeginDocument{\setlength{\wcol}{0.49\columnwidth}}
\newcommand{\widew}[1]{#1\textwidth}
\newcommand{\narroww}[1]{#1\wcol}

\title{RigidAnchor: Cross-View Geo-Localization for UAV\\ via Iterative Planar Rigid Alignment}

\author{
	\textbf{Bo Wang}$^{1}$\thanks{Email: wwcreator@qq.com} \kern 0.25em \thanks{Equal Contribution}\qquad
	\textbf{Dekang Zhu}$^{1}$\footnotemark[2]  \qquad
	\textbf{Jian Li}$^{2,3}$ \\
	\textbf{Zhipeng Meng}$^{1}$ \qquad
	\textbf{Bo Wang}$^{1}$\footnotemark[3] \qquad
	\textbf{Dewen Hu}$^{2,3}$\thanks{Corresponding authors} \\
	$^1$National Innovation Institute of Defense Technology,\\
	$^2$College of Intelligence Science and Technology, National University of Defense Technology,\\
	$^3$School of Future Technology, National University of Defense Technology
}

%% file: body/abstract.tex
\begin{abstract}
	
	A UAV that flies without GNSS has to estimate its absolute position and heading without any external signal. Most existing methods for UAV-to-satellite cross-view geo-localization (U-S CVGL) treat the task as image retrieval: they compress the UAV image and each satellite tile into separate descriptors, rank the tiles by their similarity, and return the center of the best tile. The output can therefore only lie on a pre-defined grid, its accuracy is limited by the tile size, and no heading is produced. Regressing the pose in a single pass removes the grid, but it leaves no chance to correct the estimate afterwards. We observe that satellite tiles are not independent samples. They are laid on the map at a fixed spacing, scale and orientation, so neighboring tiles join into one continuous satellite block with a consistent metric, and a nadir UAV view of that block differs from it by a planar rigid transform only, i.e. a 2D translation plus an in-plane rotation. We therefore reformulate U-S CVGL as solving for this transform between two images of different sizes. Position and heading then become parts of one geometric quantity, and the accuracy is no longer limited by the tile grid. We propose RigidAnchor, which solves the transform iteratively. The network encodes the two images and builds a multi-level correlation pyramid between them. At each iteration, it indexes the pyramid with every aerial cell carried through the current estimate, pools the descriptors of all cells into one hidden state, and lets two decoupled heads predict a translation increment and a rotation increment, the latter being composed rather than added. RigidAnchor has $4.2$M parameters and is trained with all twelve iterations supervised. On Bearing-UAV-90K, the cross-view mean errors are $78.6\%$ lower in position and $69.7\%$ lower in heading than Bearing-UAV, and HSR@15 increases by $19.06$ points. 
	Code is available at \href{https://github.com/wwsource/RigidAnchor}{https://github.com/wwsource/RigidAnchor}. 
	
\end{abstract}

%% file: body/body.tex
\section{Introduction}
A UAV that flies without a global navigation satellite system (GNSS) has to estimate its absolute position and heading without any external signal~\cite{fragoso2021seasonally, li2025geovins}. Its nadir-view camera sees the ground from above, and the geo-referenced satellite imagery that is now available for most of the world covers the same ground. The UAV view can therefore serve as an observation and the satellite image as its reference. If we register one onto the other, we can read the UAV position and heading in the map frame directly. This gives a low-cost alternative to GNSS that does not drift over time, and the task is known as UAV-to-satellite cross-view geo-localization (U-S CVGL)~\cite{zheng2020university, durgam2024cross}. U-S CVGL is useful in beyond-visual-line-of-sight inspection, target designation and terminal guidance, where the output serves as a navigation state~\cite{liu2026beyond} and as the reference for payload pointing.
Most existing methods treat U-S CVGL as image retrieval. They compress the UAV image and each satellite tile into separate feature descriptors, rank the tiles by their similarity, and take the center of the best-matching tile as the position estimate, as in NetVLAD~\cite{arandjelovic2018netvlad}, University-1652~\cite{zheng2020university} and scene-graph-based encoding-and-matching networks~\cite{duan2024scene}. Retrieval therefore turns a continuous localization problem into tile-level ranking. The output can only lie on a pre-defined regular grid, the error is limited by the tile size, and no heading is produced.
\input{fig/overview}

A second group of methods adds fine matching or direct regression to retrieval, so that the result is no longer tied to a tile center. OS-FPI~\cite{chen2024os} extracts both inputs with a single-stream network and regresses a response map, which an offset prediction then refines. The second stage of GRiM-Net~\cite{hu2026grim} detects keypoints on the candidate tiles, matches them in both directions and estimates a weighted homography. Bearing-UAV~\cite{liu2026beyond} goes further: it gives up the tile-center assumption and regresses the UAV position and heading together from the global and local structure of neighboring tiles.
Satellite tiles are not independent samples. They cover the map on a regular grid, with fixed spacing, scale and orientation, so any set of neighboring tiles forms one continuous satellite block, which covers a single geo-referenced plane with a consistent metric. A nadir UAV view observes a local part of that plane. Once the two are brought to the same ground sampling distance, the relation between them is exactly a planar rigid transform, i.e. a 2D translation plus an in-plane rotation, and not a set of independent similarity scores. Existing methods either compare the UAV image with each tile separately~\cite{arandjelovic2018netvlad, yin2023isimloc} or match points inside one candidate tile~\cite{wang2023fine, tang2025r2ploc, hu2026grim}. Both choices ignore the regularity between tiles and the global rigidity of the observed scene. If the problem is instead formulated as planar rigid transform estimation, position and orientation can be solved continuously across tiles, and the tile grid no longer limits the accuracy.

We therefore recast U-S CVGL, for the first time, as solving the planar rigid transform between two images of different sizes. Instead of scoring the UAV image against neighboring tiles one by one, we design a network that regresses the transform of the UAV image with respect to the satellite block stitched from those tiles, i.e. a 2D translation and an in-plane rotation. The network first encodes the UAV image and the satellite block into coarse features, then builds a multi-level correlation pyramid between them, and indexes the pyramid iteratively with the current estimate. At each iteration, it solves the update globally from the indexed correlation and the previous transform, and emits the rigid transform increment of that iteration in a decoupled form. Fig.~\ref{fig: overview} shows the whole pipeline.

The study offers the following contributions:
\begin{itemize}
	\item We propose RigidAnchor, which formulates U-S CVGL as solving the planar rigid transform between a UAV image and a satellite block of different sizes. This formulation puts position and heading into one geometric quantity, and it brings the accuracy below the tile level.
	\item We build a multi-level correlation pyramid between the UAV image and the satellite block, and index it dynamically at each iteration. This handles the large displacement between the initial and the ground-truth position.
	\item We solve the position and orientation update of each iteration globally and emit the rigid transform increment in a decoupled form. This improves inference efficiency. 
	\item On Bearing-UAV-90K~\cite{liu2026beyond}, in the cross-view setting, RigidAnchor reduces the mean localization error of Bearing-UAV by $78.6\%$ and its mean heading error by $69.7\%$.
\end{itemize}

\section{Related Work}
\subsection{Retrieval-Based CVGL Methods}
Cross-view geo-localization was first studied between ground panoramas and aerial imagery. CVUSA~\cite{workman2015wide} established the wide-area protocol with aerial reference imagery, CVM-Net~\cite{hu2018cvm} learned a shared embedding with a two-branch network, SAFA~\cite{shi2019spatial} introduced spatial-aware feature aggregation, and CVACT~\cite{liu2019lending} added orientation priors, so that the descriptor becomes sensitive to the relative camera azimuth. These works fixed the retrieve-and-rank paradigm: the query and every reference image are mapped to a compact descriptor, and the most similar reference gives the position.

The UAV case later received its own benchmarks and descriptors. University-1652~\cite{zheng2020university} provides a multi-view, multi-source benchmark that covers UAV, satellite and ground images, together with the Recall@K protocol. LPN~\cite{wang2022each} shows that local patterns, rather than one global vector, carry the discriminative signal. TransGeo~\cite{zhu2022transgeo} replaces the CNN backbone with a transformer to model long-range dependencies. DenseUAV~\cite{dai2024vision} moves the evaluation to low-altitude urban flight with a denser gallery, SUES-200~\cite{zhu2023sues} varies the flight altitude and the scene type, and UAV-VisLoc~\cite{xu2024uav} extends coverage to a larger geographic span. 

Later work improves the representation, the training signal or the evaluation protocol. MCCG~\cite{shen2024mccg} combines a ConvNeXt backbone with multiple classifiers, Sample4Geo~\cite{deuser2023sample4geo} shows that hard negative sampling matters greatly, VIGOR~\cite{zhu2021vigor} supplies a benchmark in which the query and the reference do not overlap one-to-one, and GeoDTR~\cite{zhang2023cross, zhang2024geodtr} disentangles the geometric layout from the appearance. PFIRNet~\cite{kang2026pfirnet} goes one step further and converts the top-$k$ retrieval responses into a probability field in coordinate space. In all of these methods the gallery is still a finite, uniformly sampled set of tiles, so the output is a tile index or a tile center, and the attainable resolution follows the sampling step rather than the network.

\subsection{Vision-Only Orientation Estimation}
Retrieval gives a position but no heading, while a navigation state needs both. Estimating the heading from images alone is therefore a problem of its own, since a magnetometer is disturbed by nearby structures and the yaw of an inertial unit drifts without bound.

The most developed route treats the task as 3-DoF ground-to-satellite localization. Shi \textit{et al.}~\cite{shi2020where} regress location and orientation together by matching a ground image against a polar-transformed satellite image. The extended journal version~\cite{shi2023accurate} refines the formulation and the training, and a geometry-guided cross-view transformer~\cite{shi2023boosting} further improves the accuracy of the estimated azimuth. SliceMatch~\cite{lentsch2023slicematch} aggregates the features into one ground descriptor and a set of pose-dependent aerial descriptors, so all candidate poses are compared at once and the pose is read from a single similarity volume.

A second route obtains the heading from an explicitly estimated geometric transform. CAHomo~\cite{wang2023fine} warps the ground image onto the satellite plane and recovers the camera orientation from a point above the central axis of the warped image, so the azimuth comes from the homography instead of from a dedicated head. A third route regresses the heading together with the position, as Bearing-UAV~\cite{liu2026beyond} does from the structural features of neighboring tiles.

\subsection{Fine-Grained Localization Methods}
Fine-grained methods replace the tile index with a metric coordinate. A common recipe matches the two images with a detector-free matcher such as LoFTR~\cite{sun2021loftr} or a sparse one such as LightGlue~\cite{lindenberger2023lightglue}, and then solves a geometric transform from the resulting correspondences.

OS-FPI~\cite{chen2024os} extracts UAV and satellite features in one stream, keeps the intermediate maps for fusion, and regresses a response map that an offset branch refines. R2PLoc~\cite{tang2025r2ploc} casts the task as region-to-point localization; it aligns multiscale semantics with a hierarchical aggregation module and enforces geometric consistency with a semantic-enhanced refinement matcher. GRiM-Net~\cite{hu2026grim} uses both granularities: it narrows the search by retrieval and then runs a keypoint-based homography estimation on the candidate patches. Bearing-UAV~\cite{liu2026beyond} skips the correspondence step and regresses position and heading from neighboring tiles, and PFIRNet~\cite{kang2026pfirnet} keeps the retrieval backbone but replaces the discrete readout with a continuous posterior.

\section{Proposed Method}
\subsection{Problem Formulation}\label{sec:problem}
The network takes two images that observe the same ground area at the same ground sampling distance (GSD). We assume that the gimbal keeps the camera pointing to the nadir, so the aerial image $\mathbf{I}^{u}\in\mathbb{R}^{S\times S\times 3}$ is a square footprint of the ground plane, and that the camera is calibrated with a known focal length in pixels. Under the pinhole model, the GSD of the aerial image and the ground side length it covers follow from the height above ground and the focal length,
\begin{equation}
	g_{u} = h/f, \qquad L = S\,g_{u},
	\label{eq:gsd}
\end{equation}
where $h$ is the height above ground, $f$ the focal length in pixels, and $L$ the ground side length covered by the aerial image.

The satellite map $\mathcal{M}$ is provided at its own GSD $g_{s}$, which is in practice coarser than $g_{u}$. We resample it to the metric resolution of the aerial image by the factor
\begin{equation}
	\alpha = g_{s}/g_{u},
	\label{eq:resample}
\end{equation}
so that one pixel of either image covers the same number of meters and no scale factor remains between them.

Given the coarse position $\mathbf{p}_{0}$ from the onboard navigation filter, we crop a satellite block $\mathbf{I}^{b}\in\mathbb{R}^{2S\times 2S\times 3}$ from the resampled map. The block is centered at $\mathbf{p}_{0}$ and covers $2L\times 2L$ on the ground, i.e. twice the side length of the aerial image.

We define a frame $\mathcal{F}_{u}$ for the aerial image, with its origin at the image center, $x$ to the right and $y$ towards the top of the image, i.e. along the nose. We define a frame $\mathcal{F}_{b}$ for the block, with its origin at the block center, $x$ to the east and $y$ to the north. Both frames are right-handed, so they differ by a rigid transform only. The quantity to estimate is the planar rigid transform $\boldsymbol{\xi}=(\tilde{\mathbf{t}},\theta)$ that maps $\mathbf{p}\in\mathcal{F}_{u}$ to $\mathbf{q}\in\mathcal{F}_{b}$,
\begin{equation}
	\mathbf{q} = R(\theta)\,\mathbf{p} + \tfrac{L}{2}\,\tilde{\mathbf{t}},
	\label{eq:rigid}
\end{equation}
where $R(\theta)\in SO(2)$ is the in-plane rotation and $\tilde{\mathbf{t}}\in\mathbb{R}^{2}$ the translation. The translation is measured in units of $L/2$: $\tilde{t}_{x}=1$ means a shift of $L/2$ to the east, and $\tilde{t}_{y}=1$ a shift of $L/2$ to the north. Because the navigation filter places $\mathbf{p}_{0}$ within $L/2$ of the true position along each axis, the ground truth satisfies $|\tilde{t}_{x}|\le 1$ and $|\tilde{t}_{y}|\le 1$. The range of $\tilde{\mathbf{t}}$ is therefore a fixed unit square, while the block itself extends to $|\tilde{t}|=2$ and thus leaves a ring of context around the aerial footprint, which the correlation stage uses. The angle $\theta$ is the rotation about the upward $z$ axis, positive counter-clockwise, so $\theta=60^{\circ}$ means that the top of the aerial image, i.e. the nose, points $60^{\circ}$ west of north.

Since $\theta$ and $\theta+2\pi$ denote the same rotation, we let the network regress the cosine and the sine of the angle rather than the angle itself, and recover $\theta=\operatorname{atan2}(\sin\theta,\cos\theta)$ at inference. This removes the discontinuity that an angular parameterization has at $\pm\pi$, and it does not restrict the range of the rotation.
\subsection{Correlation Feature Construction}\label{sec:corr}
Two branch encoders of the same architecture but with independent parameters encode the aerial image $\mathbf{I}^{u}$ and the satellite block $\mathbf{I}^{b}$. Each is a residual network of stride $8$, ending in a $1\times1$ projection to $D$ channels:
\begin{equation}
	\begin{aligned}
		\mathbf{F}^{u} &= E_{u}(\mathbf{I}^{u}) \in \mathbb{R}^{D\times s_{u}\times s_{u}}, \\
		\mathbf{F}^{b} &= E_{b}(\mathbf{I}^{b}) \in \mathbb{R}^{D\times s_{b}\times s_{b}},
	\end{aligned}
	\label{eq:feat}
\end{equation}
where $s_{u}=S/8$ and $s_{b}=S/4$. The stride is set by the cost of the next stage, since the correlation volume grows with the fourth power of the resolution, while one cell at stride $8$ still covers only $8g_{u}$ on the ground.

The two maps are correlated exhaustively. The score of an aerial cell $\mathbf{p}$ and a block cell $\mathbf{q}$ is the inner product of the two $D$-vectors,
\begin{equation}
	\mathcal{C}(\mathbf{p},\mathbf{q}) = \frac{1}{\sqrt{D}}\big\langle \mathbf{F}^{u}(\mathbf{p}),\, \mathbf{F}^{b}(\mathbf{q}) \big\rangle .
	\label{eq:corr}
\end{equation}
$\mathcal{C}$ is a four-dimensional tensor with $s_{u}^{2}s_{b}^{2}$ entries. It is computed once for each image pair, and every later access is an interpolation into it.

We pool only the block axes and organize this volume as a pyramid,
\begin{equation}
	\mathcal{C}^{1}=\mathcal{C}, \qquad
	\mathcal{C}^{l}=\operatorname{AvgPool}_{2\times 2}\big(\mathcal{C}^{l-1}\big), \quad l=2,\dots,L_{p},
	\label{eq:pyramid}
\end{equation}
so that level $l$ resolves the block at $s_{b}/2^{l-1}$ cells per side. The aerial axes stay at full resolution, because merging them would reduce the number of constraints on one global estimate.

The volume is never read at fixed positions. At iteration $k$, the current estimate $\boldsymbol{\xi}_{k}=(\tilde{\mathbf{t}}_{k},\theta_{k})$ carries every aerial cell into the block with the same rigid transform that defines the problem,
\begin{equation}
	\mathbf{q}_{k}(\mathbf{p}) = R(\theta_{k})\big(\mathbf{p}-\mathbf{c}_{u}\big) + \tfrac{s_{u}}{2}\,\tilde{\mathbf{t}}_{k},
	\label{eq:index}
\end{equation}
where $\mathbf{c}_{u}=(s_{u}-1)/2$ is the center of the aerial lattice and the factor $s_{u}/2$ is the feature-level image of $L/2$. At each level, a $(2r+1)\times(2r+1)$ window is read around the query position $\mathbf{r}^{l}_{k}(\mathbf{p})=2^{1-l}\,(\mathbf{c}_{b}+\mathbf{q}_{k}(\mathbf{p}))$ by bilinear interpolation, and the windows are concatenated into one descriptor per cell,
\begin{equation}
	\mathbf{z}_{k}(\mathbf{p}) = \operatorname*{concat}_{\,l,\,\mathbf{d}}\ \mathcal{C}^{l}\big(\mathbf{r}^{l}_{k}(\mathbf{p}) + \mathbf{d}\big),
	\label{eq:indexfeat}
\end{equation}
where $\mathbf{c}_{b}=(s_{b}-1)/2$, the level runs over $l=1,\dots,L_{p}$ and $\mathbf{d}$ over the integer offsets of the window. The descriptor therefore has $L_{p}(2r+1)^{2}$ values, i.e. $324$ with $L_{p}=r=4$. Because the volume is built once and only re-read, one iteration costs a set of interpolations whatever the search range, and the $s_{u}^{2}$ descriptors all describe the same rigid transform, which the next stage reduces to one global update.
\subsection{Iterative Pose Update}\label{sec:update}
The update acts on all $s_{u}^{2}$ aerial cells at once (Fig.~\ref{fig: update}). A context encoder $E_{c}$ is applied to the aerial image and pooled to a single point, which gives the initial hidden state $\mathbf{h}_{0}\in\mathbb{R}^{128}$ and a context vector $\mathbf{g}\in\mathbb{R}^{128}$. At iteration $k$, the descriptors of all cells go through a motion encoder and are averaged into one vector $\mathbf{m}_{k}=\rho(\{\mathbf{z}_{k}(\mathbf{p})\}_{\mathbf{p}})$, where $\rho$ denotes the encoder followed by average pooling. Together with the current estimate, it drives a GRU,
\begin{equation}
	\mathbf{h}_{k} = \mathrm{GRU}\big(\mathbf{h}_{k-1},\, [\mathbf{m}_{k};\, \mathbf{g};\, \cos\theta_{k-1};\, \sin\theta_{k-1};\, \tilde{\mathbf{t}}_{k-1}]\big),
	\label{eq:gru}
\end{equation}
Because the pooling runs over every cell, one hidden state summarizes the whole footprint, and an iteration returns a single update of the rigid transform instead of a displacement per cell. No flow field is decoded, and no transform is fitted afterwards.
\input{fig/update}

Two heads $H_{t}$ and $H_{a}$ read the same hidden state,
\begin{equation}
	\Delta\tilde{\mathbf{t}}_{k} = H_{t}(\mathbf{h}_{k}), \qquad
	(\Delta c_{k},\,\Delta s_{k}) = H_{a}(\mathbf{h}_{k}),
	\label{eq:heads}
\end{equation}
The two parts of $\boldsymbol{\xi}$ are then updated separately,
\begin{equation}
	\tilde{\mathbf{t}}_{k} = \tilde{\mathbf{t}}_{k-1} + \Delta\tilde{\mathbf{t}}_{k}, \qquad
	R(\theta_{k}) = R(\Delta\theta_{k})\,R(\theta_{k-1}),
	\label{eq:update}
\end{equation}
where the rotation is evaluated as $\cos\theta_{k}=\Delta c_{k}\cos\theta_{k-1}-\Delta s_{k}\sin\theta_{k-1}$ and $\sin\theta_{k}=\Delta s_{k}\cos\theta_{k-1}+\Delta c_{k}\sin\theta_{k-1}$. The rotation is composed rather than added, so no angle enters the network. Since the two heads are separate, a translation error is not expressed as a spurious rotation, and vice versa. The predicted pair is not constrained to unit norm, so the composed operator carries the product of the norms of the increments. That norm, written $|R|$ in what follows, also sets the scale of the search window through which the correlation volume is read. We therefore renormalize $(\cos\theta_{k},\sin\theta_{k})$ to unit length after each composition, which keeps $R(\theta_{k})$ an orthogonal matrix and the window scale fixed. Every iteration emits a complete estimate $\boldsymbol{\xi}_{k}$, which is read out through Eq.~\eqref{eq:rigid}.
\subsection{Training and Inference}\label{sec:infer}
Let $K$ be the number of iterations. Every iteration emits a complete estimate, and all of them are supervised,
\begin{equation}
	\mathcal{L} = w_{p}\sum_{k=1}^{K}\gamma^{K-k}\,\ell\big(\tilde{\mathbf{t}}_{k}-\tilde{\mathbf{t}}^{*}\big)
	+ (1-w_{p})\sum_{k=1}^{K}\gamma^{K-k}\,\ell\big(\mathbf{w}_{k}-\mathbf{w}^{*}\big),
	\label{eq:loss}
\end{equation}
where $\mathbf{w}_{k}=(\cos\theta_{k},\sin\theta_{k})$ is the predicted rotation and $\mathbf{w}^{*}$ its ground truth, $\tilde{\mathbf{t}}^{*}$ is the ground-truth translation in the same normalized unit, $\ell$ is the smooth-$L_{1}$ norm, $w_{p}$ is the weight of the position term and $1-w_{p}$ that of the direction term, and $\gamma=0.8$ with $w_{p}=0.5$ in all reported results. Since the translation is measured in units of $L/2$ and the rotation is a unit vector, both targets are dimensionless and of order one. One loss and one weight therefore serve both, and no per-quantity rescaling is needed. The weight $\gamma^{K-k}$ is largest for the last iteration and decays towards the earlier ones, because only $\boldsymbol{\xi}_{K}$ is deployed, while every earlier estimate still has to improve on its predecessor. Supervising the pair $(\cos\theta,\sin\theta)$ instead of an angle keeps this term continuous across $\pm\pi$.

At inference the estimate starts from $\boldsymbol{\xi}_{0}=(\mathbf{0},0)$, i.e. from the coarse position with the nose pointing north. Since the iterations predict increments, this choice only sets where the sampling grid starts, and the error it leaves is what the following iterations are trained to remove. We run $K=12$ iterations, the same as in training, and read out $\hat{\boldsymbol{\xi}}=\boldsymbol{\xi}_{K}$ through Eq.~\eqref{eq:rigid}. Because every iteration is already a complete estimate, $K$ can be shortened at test time to trade accuracy for latency. One forward pass therefore returns both components of a navigation state, position and heading.

\section{Experiments}
\subsection{Experimental Setup}
\subsubsection{Dataset}
We evaluate RigidAnchor on Bearing-UAV-90K~\cite{liu2026beyond}, a cross-view benchmark built from contiguous satellite imagery of four cities. Its UAV views are not aligned with the satellite tiles, and each view carries a heading annotation. Each city contributes one orthoimage of $4096\times4096$ pixels at $0.25$~m/px, which is cut into $16\times16$ tiles of $256\times256$ pixels, i.e.\ $64$~m on the ground. Every $2\times2$ group of adjacent tiles forms one block, so a city yields $15\times15$ blocks. The four cities therefore give $1024$ tiles and $900$ blocks in total. Inside each block, $100$ camera poses are drawn with a random position and a random yaw, and the corresponding views are rendered from the photogrammetric 3D model of Google Earth. A UAV view therefore differs from its geo-referenced reference by an arbitrary rotation and an arbitrary offset, rather than by a tile index, which is exactly the setting that Sec.~\ref{sec:problem} formulates. The four cities provide 90K UAV views in total.

Each sample provides the four tiles adjacent to the sampled position, which we stitch into the $512\times512$ block $\mathbf{I}^{b}$, together with the annotation pair $(\tilde{\mathbf{t}}^{*},\mathbf{w}^{*})$ used in Eq.~\eqref{eq:loss}: the position of the view in the frame of the block center, in units of $L/2$, and the heading as the unit vector $(\cos\theta^{*},\sin\theta^{*})$. The block center plays the role of the coarse position $\mathbf{p}_{0}$, so the annotation is exactly the quantity that the network estimates, and it lies inside the unit square of Sec.~\ref{sec:problem} by construction. The tiles are already provided at $0.25$~m/px, i.e.\ $g_{s}=g_{u}$, so $\alpha=1$ in Eq.~\eqref{eq:resample} and no resampling artifact enters the comparison.

The 90K samples are split at random, with a fixed seed, into $85\%$, $5\%$ and $10\%$ for training, validation and testing.
\input{table/loc-orientation}
\subsubsection{Implementation Details}
We implement RigidAnchor in PyTorch and run all experiments on a single NVIDIA RTX 4090 GPU. The aerial image is $256\times256$ pixels and the satellite block $512\times512$, which gives feature maps of $32\times32$ and $64\times64$ cells at stride $8$. Both branch encoders are trained from scratch, and no photometric augmentation is used. Images are normalized by the ImageNet mean and standard deviation before encoding. We set $D=256$, $L_{p}=4$ and $r=4$, so the descriptor of Eq.~\eqref{eq:indexfeat} has $324$ channels, and we use $128$ channels for the GRU and for the context encoder $E_{c}$, with $K=12$ iterations. The whole network has $4.2$M parameters: the two branch encoders account for $2.1$M, the context encoder for about $1.1$M, and the update module for about $1.0$M.
\subsubsection{Training Protocol}
Training has a single stage. It starts from random initialization and minimizes Eq.~\eqref{eq:loss} on the training split, where all $K=12$ iterations are supervised with $\gamma=0.8$ and $w_{p}=0.5$. We use AdamW with weight decay $10^{-5}$ and $\epsilon=10^{-8}$, a batch size of $16$, gradient clipping at norm $1.0$ and mixed precision, for $40$ epochs. The learning rate follows a one-cycle schedule: it rises to $2\times10^{-4}$ within the first $1.5\%$ of the steps and then decays linearly to zero. We evaluate the validation split every $4$ epochs and use the weights of the last epoch for all results below. At test time the iterations start from $\boldsymbol{\xi}_{0}=(\mathbf{0},0)$, run the same $12$ steps, and the last estimate is read out.
\subsection{Benchmark Results on Bearing-UAV-90K: Localization and Orientation}
Table~\ref{tab: loc-orientation} compares RigidAnchor with four retrieval-based methods and with Bearing-UAV on Bearing-UAV-90K. In the Sat. setting the query is a crop of the satellite imagery itself, so this setting measures how finely a method puts an image back on its own source. In the UAV setting the query is a rendered view with an arbitrary heading and offset, which is the cross-view setting of Sec.~\ref{sec:problem}. All methods are evaluated on the same test split. The two success rates use one band each: a query counts for LSR@15 when its position error is within $15$~m, and for HSR@15 when its heading error is within $15^{\circ}$.

The four retrieval-based methods rank the gallery correctly but locate coarsely. Recall@1 ranges from $60.20$ to $80.04$, yet LSR@15 stays between $15.11$ and $31.33$. One tile covers $64$~m on the ground, so even the correct tile leaves its center tens of meters away. A tile index also carries no heading, which is why these rows have neither HSR@15 nor MHE. GTA-UAV adds a coordinate regression on top of the retrieval and does better, but its mean error of $28.43$~m is still about twice the tolerance and its LSR@15 only reaches $27.96$. Bearing-UAV gives up the tile-center assumption and reaches $98.33$ and $98.12$ on the Sat. side, but falls to $89.36$ and $77.21$ with an MHE of $12.90^{\circ}$ on the UAV side. A single regression pass leaves no chance to correct the estimate, so the difference in appearance between the two views remains in the result.

RigidAnchor leads all ten columns: LSR@15 $100.00$ and $97.53$, HSR@15 $100.00$ and $96.27$, Recall@1 $99.79$ and $96.31$, with mean errors of $0.10$~m and $1.84$~m in position and $0.07^{\circ}$ and $3.91^{\circ}$ in heading. Against Bearing-UAV, MLE falls by $98.2\%$ and $78.6\%$, MHE by $98.3\%$ and $69.7\%$, and HSR@15 rises by $1.88$ and $19.06$ points. On the Sat. side both LSR@15 and HSR@15 saturate, so no query leaves the $15$~m or $15^{\circ}$ band. The remaining $0.10$~m is $0.4$ pixel, far below the $2$~m of one stride-$8$ cell, so the residual is no longer set by the search range or by the feature grid.

The two settings differ more in relative terms only because the Sat. denominator is close to zero. In absolute terms the UAV errors are $12\%$ and $26\%$ of the tolerances, so the estimate can serve directly as a navigation state.

Fig.~\ref{fig: error} shows the UAV-view errors sample by sample over the same $9000$ queries. The median is $0.84$~m and $1.26^{\circ}$ for RigidAnchor, against $7.23$~m and $7.12^{\circ}$ for Bearing-UAV, so half of our queries are placed within $3.4$ pixels and oriented within $1.3^{\circ}$, and $90\%$ within $2.19$~m and $4.22^{\circ}$, against $15.46$~m and $25.69^{\circ}$. In both rows the mean lies above the median, in ours by more than a factor of two, so the remaining error comes from a minority of samples rather than from a uniform bias. It is largely the same minority in both rows: of the $984$ queries that Bearing-UAV places beyond $15$~m, we bring $794$ ($81\%$) inside, at a median of $1.06$~m, and only $32$ of the $8016$ it does place inside leave the band. The per-sample panels show that this minority is not confined to any range of indices, and the maximum distance is still $77.1$~m, against $77.2$~m for Bearing-UAV, so the hardest queries are the ones the method does not fix.
\subsection{Ablation Studies}
\input{fig/error}

\input{table/design}
\subsubsection{Key Design Choices}
Table~\ref{tab: design} varies the four design choices that the method leaves open. Each of them is a switch: the position weight $w_p$ of Eq.~\eqref{eq:loss}, the direction term taking $1-w_p$; whether the satellite block is encoded by its own branch $E_b$ or by the branch that encodes the UAV view; whether the images are normalized by the ImageNet statistics before encoding; and whether the predicted rotation is renormalized to unit length after each composition (Sec.~\ref{sec:update}). Any two consecutive rows differ in exactly one switch, so the four adjacent pairs give a direct reading of what that switch is worth. The last row is the complete model.

No single switch dominates, and they do not act on the same error channel. Lowering $w_p$ from $0.8$ to $0.5$, i.e.\ raising the weight of the direction term from $0.2$ to $0.5$, moves the mean errors from $2.55$~m and $5.35^{\circ}$ to $2.38$~m and $4.63^{\circ}$. In relative terms the heading error falls by $13\%$ but the position error by only $6.7\%$, and $13\%$ is the largest heading gain of any switch, since a $4{:}1$ weighting leaves the direction term the most to gain. The separate branch $E_b$ acts the other way: it is worth $8.4\%$ of the position error but only $6.9\%$ of the heading error, so what the shared branch costs is mostly positional. ImageNet normalization moves the two means by only $0.13$~m and $0.08^{\circ}$, yet it lifts Recall@1 by $0.44$ points. It changes only the statistics of the input, not the architecture or the loss. It should therefore act on whether the two views are matched at all, rather than on how precisely they are aligned, and the two readings say exactly that. Renormalizing the rotation gives the largest positional gain of the four, $0.21$~m or $10\%$, with the heading error down by $0.32^{\circ}$ or $7.6\%$. The other three switches require retraining, since they change the loss, the encoders or the input statistics. This one acts at inference only, where it restores the composed operator to unit norm after each composition, so it needs no retraining at all. Sec.~\ref{sec:update} explains why the composed operator drifts in scale, and Fig.~\ref{fig: iters} plots the same effect against the iteration count.

Together the four switches lower the mean errors from $2.55$~m to $1.84$~m ($27.8\%$) and from $5.35^{\circ}$ to $3.91^{\circ}$ ($26.9\%$), and the five configurations stay within a band of $0.71$~m and $1.44^{\circ}$. That band is much narrower than the distance to Bearing-UAV ($8.61$~m and $12.90^{\circ}$), so the accuracy comes from the formulation of Sec.~\ref{sec:problem} and the iterative solution of Secs.~\ref{sec:corr} and~\ref{sec:update}, not from these choices. Each of them recovers between six and ten percent of the position error on its own.
\subsubsection{Grid Point Density}
\input{table/grid}
\input{fig/iters}
Table~\ref{tab: grid} asks how many aerial cells have to vote for the single motion vector $\mathbf{m}_{k}$ that drives the update. In the model of Sec.~\ref{sec:update}, all $s_{u}^{2}=1024$ descriptors are encoded and averaged; here only $n\times n$ of them are, evenly spaced over the lattice and always including the four corners, so $n=2$ keeps four cells, i.e.\ $0.4\%$ of the footprint. One model is trained per density, and each is evaluated at its own setting. Both errors fall monotonically with the number of cells, from $7.23$~m and $12.47^{\circ}$ at $n=2$ to $1.84$~m and $3.91^{\circ}$ at $n=32$, i.e.\ by a factor of $3.9$ in position and $3.2$ in heading. Recall@1, which only records whether the returned position falls in the correct tile, changes much less, from $85.50$ to $96.31$: the density governs the fine alignment, not the coarse one.

The gain, however, comes early. From four cells to $16$, the two errors drop by $3.80$~m and $6.82^{\circ}$, which is $71\%$ and $80\%$ of the total. By $64$ cells they have reached $94\%$ and $95\%$, and the last two settings together add only $0.33$~m and $0.42^{\circ}$. A quarter of the lattice, $n=16$, thus keeps $95\%$ of the position reduction and $97\%$ of the heading one.

The $n=2$ row also shows where the formulation, and not the density, does the work. Four corner cells already reach $85.50\%$ Recall@1 and $7.23$~m, both better than the $83.17\%$ and $8.61$~m of Bearing-UAV, so even this minimal setting localizes. What four cells cannot do is refine, and the reason is geometric. A residual rotation $\Delta\theta$ displaces a cell at radius $r$ from the center by $r\Delta\theta$, so the four corners, at $1.41$ times the radius of a mid-edge cell, are the cells whose descriptors the residual rotation corrupts most. An average over four values also keeps the fluctuation between cells, which an average over $1024$ would suppress. This confirms quantitatively the argument of Sec.~\ref{sec:corr}, that the aerial axes must stay at full resolution because merging them would reduce the number of constraints on one global estimate. We keep $n=32$: the convolutions that produce the correlation features run over the full lattice whatever $n$ is, so subsampling saves nothing, and the density is a question of redundancy rather than of cost.
\subsubsection{Number of Iterations}
Fig.~\ref{fig: iters} sweeps the number of refinement iterations at inference from $K=1$ to $K=32$ with the weights fixed, over the $9000$ test queries of the UAV-view setting, and it repeats the sweep with and without the rotation normalization of Sec.~\ref{sec:update}. That normalization is an identity at $|R|=1$, so both conditions start from the same first estimate, $10.86$~m and $10.31^{\circ}$. On the normalized condition, which is the one we deploy, $76.7\%$ of the total reduction in position error is obtained within two iterations and $97.4\%$ within four. The recall within $5$~m rises from $22.7\%$ at one iteration to $94.7\%$ at four and $95.8\%$ at twelve, and from eight iterations on the curve is flat, at $1.852$, $1.843$ and $1.851$~m for $8$, $12$ and $32$. The training setting of twelve iterations is therefore not what limits the accuracy.
\input{fig/orient}
\input{fig/convergence}
The two conditions diverge from the second iteration on, and the gap widens from $1.1\%$ at that iteration to $10\%$ at twelve. However, what the normalization prevents appears only after the loop has converged, for the reason given in Sec.~\ref{sec:update}. Translation is accumulated additively, so a further iteration only trims the residual. The rotation is composed through a predicted pair whose norm is not normalized, so its scale is multiplied at every step, and the window through which the correlation volume is read shrinks with it. Without normalization the position error bottoms out at $2.015$~m at iteration $8$ and returns to $2.092$~m at $32$, and the heading error rises from $4.177^{\circ}$ to $4.760^{\circ}$. With it the position error reaches $1.842$~m at iteration $14$ and stays within $0.01$~m of that value through $32$, and the heading stays within $0.04^{\circ}$ of its own optimum. Pairing the two conditions sample by sample over the same $9000$ queries, normalization lowers the position error at twelve iterations by $0.206$~m on average and the heading error by $0.319^{\circ}$; at $32$ iterations the two gains reach $0.240$~m and $0.808^{\circ}$. We keep $K=12$ to match training. Four iterations already recover $94.7\%$ of the $5$~m recall, so a shorter schedule can be used when latency matters more than accuracy.
\subsubsection{Impact of the Orientation Gap}
The orientation gap of a query is the angle $|\theta^{*}|$ between the heading of the UAV view and the orientation of the map-aligned satellite block. Bearing-UAV-90K draws it uniformly rather than concentrating it near zero. The standard deviation of $\theta^{*}$ is $103.95^{\circ}$ over the $76\,500$ training views and $104.37^{\circ}$ over the $9000$ test views, against $103.92^{\circ}$ for a uniform law, and the Rayleigh, Kuiper and $\chi^{2}$ tests all return $p>0.18$, so none of the three rejects uniformity. Gaps of any size are therefore neither scarce nor over-represented, and no re-weighting of the test set is needed to read Fig.~\ref{fig: orient}.

Fig.~\ref{fig: orient} asks the question that matters at deployment: whether the size of the gap changes the chance that a query meets the $5^{\circ}$ and $2$~m tolerance at the returned estimate. It does not. Across the twelve bins the rate ranges from $83.5\%$ to $87.4\%$ around the overall $85.9\%$, no bin departs from that level by more than $2.4$ points, and every Wilson interval covers it. The spread between bins is therefore what the roughly $750$ queries of each bin produce by sampling, not an effect of the gap. Pooling the two outer bins on each side, where a penalty would concentrate if there were one, gives $85.4\%$ for $|\theta^{*}|\leq30^{\circ}$ against $85.3\%$ for $|\theta^{*}|\geq150^{\circ}$. The $85.9\%$ we report is therefore not carried by the queries that happen to be nearly map-aligned. Since the gap itself is uniform over the test set, that rate holds for a UAV that enters the block at any heading.
\subsection{Convergence Visualization}
Fig.~\ref{fig: convergence} draws the estimate at successive iterations of a single forward pass. Each row is one query: the UAV view on the left, followed by the same $512\times512$ satellite block, on which the footprints of iterations $1$, $2$, $3$ and $12$ accumulate across the four panels. The footprint is read out of the network rather than drawn for the figure. We place a square of the same side length as the aerial coverage at the current $\tilde{\mathbf{t}}_{k}$ and rotate it by the current $\theta_{k}$, so its side spans exactly half of the block, and its offset and tilt with respect to the ground-truth footprint are the two errors of that iteration. The side length never changes, so every panel shows a purely rigid motion, which is what the two offsets and one angle of $\boldsymbol{\xi}$ predict. The three queries are not easy starts: their initial errors, $8.26$, $12.74$ and $23.03$~m, range from below to above the $9.17$~m median of the first iteration over the split.

The two components converge on different schedules. For the first query the localization error falls from $8.26$~m to $1.87$, $0.60$ and $0.27$~m, and the heading error from $11.23^{\circ}$ to $6.70^{\circ}$, $3.94^{\circ}$ and $0.10^{\circ}$; for the second, from $12.74$~m and $12.43^{\circ}$ to $3.49$~m and $11.27^{\circ}$, $0.75$~m and $4.46^{\circ}$, $0.08$~m and $0.55^{\circ}$; for the third, from $23.03$~m and $6.37^{\circ}$ to $2.31$~m and $5.35^{\circ}$, $1.22$~m and $0.45^{\circ}$, $0.39$~m and $0.02^{\circ}$. Position is finished early in every case: by the third iteration it has covered $96$, $95$ and $96\%$ of its total reduction, and the nine remaining iterations add only four or five percent. Heading is the slower channel in the first two, at $66$ and $67\%$ of its reduction by the third iteration, whereas the third query starts with the smallest heading error and has already covered $93\%$. Fig.~\ref{fig: iters} shows the same course over the $9000$ queries, where $97\%$ of the position gain is reached at the fourth iteration; these rows are its per-sample picture. 
\section{Conclusion}

In this paper, we recast UAV-to-satellite cross-view geo-localization as solving the planar rigid transform between two images of different sizes, and we propose RigidAnchor to solve it iteratively. Every iteration returns a complete estimate, position and heading come from one geometric quantity, and the accuracy is no longer set by the tile size. On Bearing-UAV-90K the cross-view mean errors are $78.6\%$ lower in position and $69.7\%$ lower in heading than Bearing-UAV, the median over the $9000$ test queries is $0.84$~m and $1.26^{\circ}$, and $85.9\%$ of the queries meet $5^{\circ}$ and $2$~m. Each of the four design switches is worth six to ten percent of the position error, and the five configurations stay within a band far narrower than the gap to Bearing-UAV, so the accuracy comes from the formulation and the iterative solution rather than from these choices. The success rate does not change detectably with the orientation gap, and the first three iterations already carry $96\%$ of the position reduction in the cases we visualize.

This work also shows that the regularity of the map is itself a usable prior. If tiles are treated as independent samples, a method can only rank them. Once neighboring tiles are seen as contiguous samples of one plane, localization becomes continuous registration, and the heading follows as a by-product instead of as an extra task. RigidAnchor still starts from a coarse position given by the navigation filter, and it assumes a nadir camera of known focal length over flat ground, so an error in height enters the scale through the ground sampling distance. Folding this uncertainty, the camera tilt and the terrain relief into the same estimate, and coupling the iteration tightly with inertial navigation, are the directions we are pursuing.

%% file: fig/overview.tex
%

\begin{figure*}[t]
\begin{center}
\includegraphics[width=\widew{1.0}]{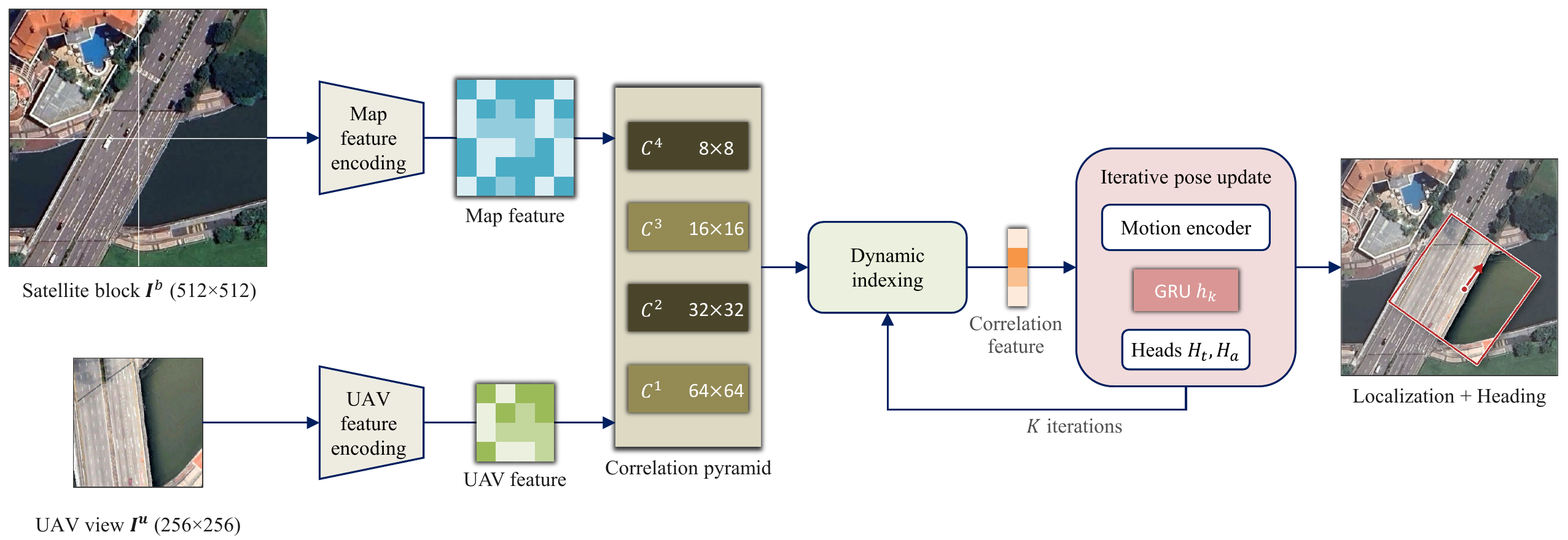}
\end{center}
\caption{
	\textbf{Overview of RigidAnchor.}
	A UAV view $\mathbf{I}^{u}$ and the satellite block $\mathbf{I}^{b}$ that contains it are encoded by two independent branches and correlated into a four-level pyramid $\mathcal{C}^{1},\dots,\mathcal{C}^{4}$, so that one volume serves every candidate pose. At each iteration the pyramid is indexed at the current estimate, the descriptors of all aerial cells are reduced by a motion encoder and a GRU to one hidden state $\mathbf{h}_{k}$, and two heads predict a single rigid increment. After $K$ iterations the estimate is read out as localization and heading.
}
\label{fig: overview}
\end{figure*}

%% file: fig/update.tex
%

\begin{figure}[t]
\begin{center}
\includegraphics[width=\narroww{1.0}]{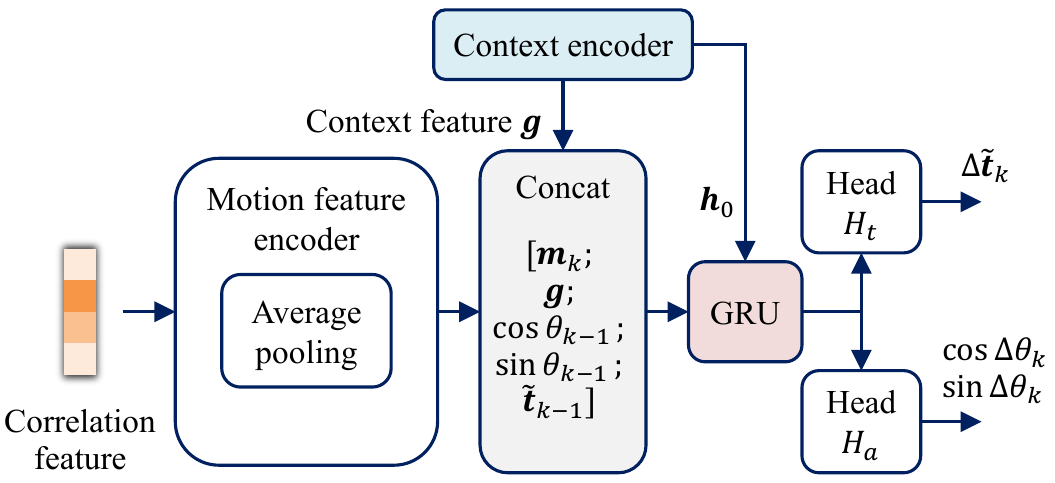}
\end{center}
\caption{
	\textbf{One iteration of the pose update.}
	The correlation features of all $s_{u}^{2}$ aerial cells are encoded and average-pooled into one motion vector $\mathbf{m}_{k}$. It is concatenated with the context vector $\mathbf{g}$ and the current estimate $(\cos\theta_{k-1},\sin\theta_{k-1},\tilde{\mathbf{t}}_{k-1})$ to drive a GRU from the hidden state $\mathbf{h}_{0}$ given by the context encoder. Two separate heads read the resulting state $\mathbf{h}_{k}$: one predicts the translation increment $\Delta\tilde{\mathbf{t}}_{k}$, and the other the rotation increment as the pair $(\cos\Delta\theta_{k},\sin\Delta\theta_{k})$, which is composed with the previous rotation instead of being added to an angle.
}
\label{fig: update}
\end{figure}

%% file: table/loc-orientation.tex
%
%

\begin{table*}[t]
\centering
\setlength{\tabcolsep}{4.6pt}
\caption{Localization and orientation results on Bearing-UAV-90K under the satellite-view (Sat.) and UAV-view (UAV) settings. Recall@1 is the fraction of queries whose returned position falls in the correct tile, LSR@15 and HSR@15 are the success rates within $15$\,m and $15^{\circ}$, and MLE and MHE the mean localization and heading errors. Best results are in boldface, and ``--'' marks a metric that the method does not produce.}
\label{tab: loc-orientation}
\resizebox{\widew{0.8}}{!}{%
\begin{tabular}{@{}l *{10}{c}@{}}
\toprule
& \multicolumn{2}{c}{Recall@1 (\%) $\uparrow$}
& \multicolumn{2}{c}{LSR@15 (\%) $\uparrow$}
& \multicolumn{2}{c}{HSR@15 (\%) $\uparrow$}
& \multicolumn{2}{c}{MLE (m) $\downarrow$}
& \multicolumn{2}{c}{MHE ($^{\circ}$) $\downarrow$} \\
\cmidrule(lr){2-3} \cmidrule(lr){4-5} \cmidrule(lr){6-7} \cmidrule(lr){8-9} \cmidrule(lr){10-11}
Method & Sat. & UAV & Sat. & UAV & Sat. & UAV & Sat. & UAV & Sat. & UAV \\ \midrule
University-1652~\cite{zheng2020university}
    & 63.29 & 60.20
    & 16.31 & 15.11
    & -- & --
    & 31.78 & 33.15
    & -- & -- \\
SUES-200~\cite{zhu2023sues}
    & 71.96 & 66.60
    & 16.78 & 15.76
    & -- & --
    & 29.09 & 30.83
    & -- & -- \\
DenseUAV~\cite{dai2024vision}
    & 80.04 & 73.43
    & 17.23 & 16.54
    & -- & --
    & 26.82 & 28.79
    & -- & -- \\
GTA-UAV~\cite{ji2025game4loc}
    & 76.99 & 70.71
    & 31.33 & 27.96
    & -- & --
    & 25.24 & 28.43
    & -- & -- \\ \midrule
Bearing-UAV~\cite{liu2026beyond}
    & 90.76 & 83.17
    & 98.33 & 89.36
    & 98.12 & 77.21
    & 5.66 & 8.61
    & 4.15 & 12.90 \\ \midrule
Ours
    & \textbf{99.79} & \textbf{96.31}
    & \textbf{100.00} & \textbf{97.53}
    & \textbf{100.00} & \textbf{96.27}
    & \textbf{0.10} & \textbf{1.84}
    & \textbf{0.07} & \textbf{3.91} \\ \bottomrule
\end{tabular}%
}
\end{table*}

%% file: fig/error.tex
%

\begin{figure*}[t]
\begin{center}
\includegraphics[width=\widew{1.0}]{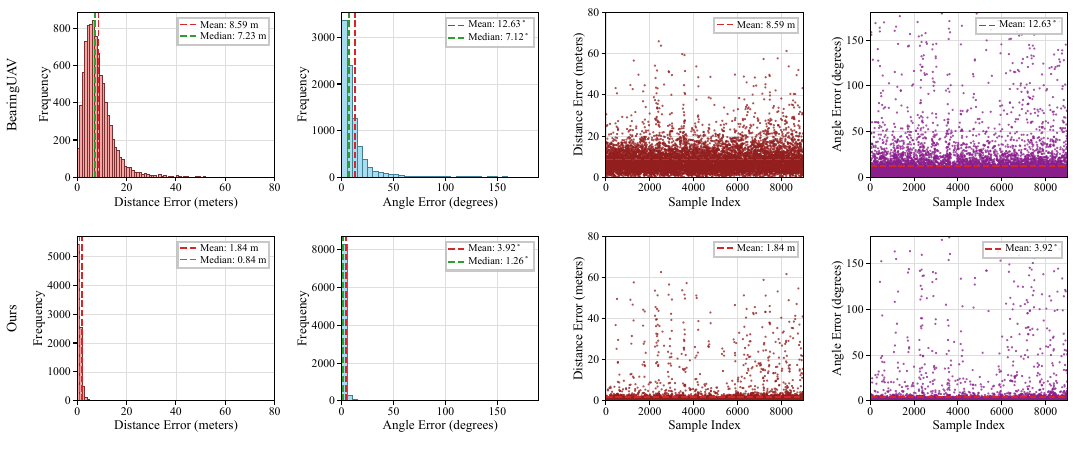}
\end{center}
\caption{
	\textbf{Error distributions on Bearing-UAV-90K.}
	Errors of Bearing-UAV (top) and RigidAnchor (bottom) over the $9000$ queries of the UAV-view setting. From left to right: histograms of the localization error and of the heading error, and the per-sample localization and heading errors. Dashed and dash-dotted lines mark the mean and the median. The Bearing-UAV row comes from a separate run of that baseline under the same protocol, so its mean errors differ slightly from the Bearing-UAV row of Table~\ref{tab: loc-orientation}.
}
\label{fig: error}
\end{figure*}

%% file: table/design.tex
\begin{table*}[t]
\centering
\setlength{\tabcolsep}{8pt}
\caption{Ablation on the key design choices, on the UAV-view test split of Bearing-UAV-90K ($9000$ queries). Each row is fixed by the four switches, and the last row is the complete model; any two consecutive rows differ in exactly one switch. The weight $w_p$ applies to the position sum of Eq.~\eqref{eq:loss} and $1-w_p$ to the direction sum, $E_b$ is the branch that encodes the satellite block of Sec.~\ref{sec:corr}, and $|R|$ norm.\ renormalizes the predicted rotation to unit length after every composition (Sec.~\ref{sec:update}). Best results are in boldface.}
\label{tab: design}
\resizebox{\widew{0.9}}{!}{%
\begin{tabular}{@{}ccccccc@{}}
\toprule
Pos.\ loss weight $w_p$ & Sep.\ enc.\ $E_b$ & ImageNet norm. & $|R|$ norm.
& \shortstack{Recall@1\\$(\%)\uparrow$}
& \shortstack{MLE\\$(m)\downarrow$}
& \shortstack{MHE\\$(^{\circ})\downarrow$} \\
\midrule
$0.8$ & $\boldsymbol{\times}$ & $\boldsymbol{\times}$ & $\boldsymbol{\times}$
    & $94.64$ & $2.55$ & $5.35$ \\
$0.5$ & $\boldsymbol{\times}$ & $\boldsymbol{\times}$ & $\boldsymbol{\times}$
    & $94.83$ & $2.38$ & $4.63$ \\
$0.5$ & \checkmark & $\boldsymbol{\times}$ & $\boldsymbol{\times}$
    & $95.52$ & $2.18$ & $4.31$ \\
$0.5$ & \checkmark & \checkmark & $\boldsymbol{\times}$
    & $95.96$ & $2.05$ & $4.23$ \\
$0.5$ & \checkmark & \checkmark & \checkmark
    & $\mathbf{96.31}$ & $\mathbf{1.84}$ & $\mathbf{3.91}$ \\
\bottomrule
\end{tabular}}
\end{table*}

%% file: table/grid.tex
%
%
%
%
%

\begin{table}[t]
\centering
\setlength{\tabcolsep}{6pt}
\caption{Ablation on the grid point density, on the UAV-view test split of Bearing-UAV-90K ($9000$ queries). Only $n\times n$ of the $32\times32$ aerial cells enter the average that forms the motion vector $\mathbf{m}_{k}$ of Sec.~\ref{sec:update}. The retained cells are evenly spaced, and the four corners are always kept, so $n=32$ is the full lattice used everywhere else. Each density is trained and evaluated at its own setting. Best results are in boldface.}
\label{tab: grid}
\resizebox{\narroww{0.77}}{!}{%
\begin{tabular}{@{}lccccc@{}}
\toprule
Grid $n$ & Cells
& \shortstack{Recall@1\\$(\%)\uparrow$}
& \shortstack{MLE\\$(m)\downarrow$}
& \shortstack{MHE\\$(^{\circ})\downarrow$} \\
\midrule
$2$  & $4$    & $85.50$ & $7.23$ & $12.47$ \\
$4$  & $16$   & $92.94$ & $3.43$ & $5.65$ \\
$8$  & $64$   & $95.59$ & $2.17$ & $4.33$ \\
$16$ & $256$  & $95.64$ & $2.11$ & $4.19$ \\
$32$ & $1024$ & $\mathbf{96.31}$ & $\mathbf{1.84}$ & $\mathbf{3.91}$ \\
\bottomrule
\end{tabular}}
\end{table}

%% file: fig/iters.tex
%

\begin{figure*}[t]
\begin{center}
\includegraphics[width=\widew{1.0}]{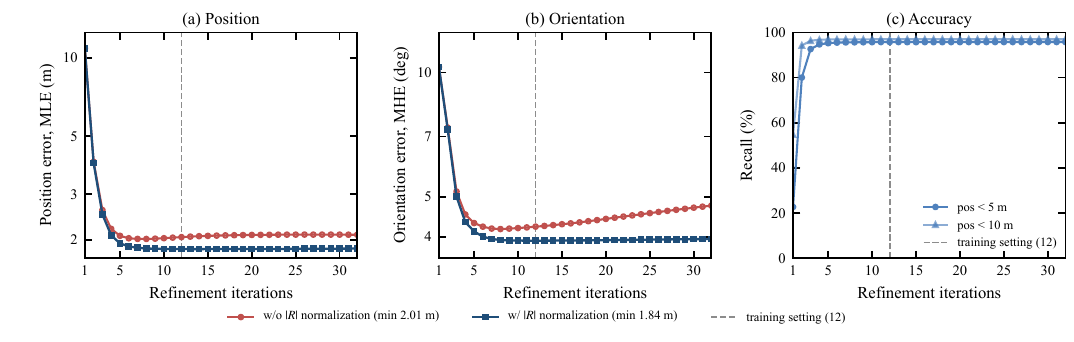}
\end{center}
\caption{
	\textbf{Number of refinement iterations, with and without rotation normalization.}
	On the UAV-view split of Bearing-UAV-90K ($9000$ queries) the same weights are run with $K$ swept from $1$ to $32$: (a) localization and (b) heading error on a logarithmic axis, with (blue) and without (red) renormalizing the rotation to unit length after each update, and (c) the recall of the normalized variant at $5$ and $10$~m. Both conditions share the first iteration, where the normalization is an identity, and they separate from the second on. Without it the error bottoms out at iteration $8$ and then rises; with it the minimum ($1.84$ against $2.01$~m) holds through $32$. The dashed gray line marks the training setting $K=12$.
}
\label{fig: iters}
\end{figure*}

%% file: fig/orient.tex
%

\begin{figure*}[t]
\begin{center}
\includegraphics[width=\widew{1.0}]{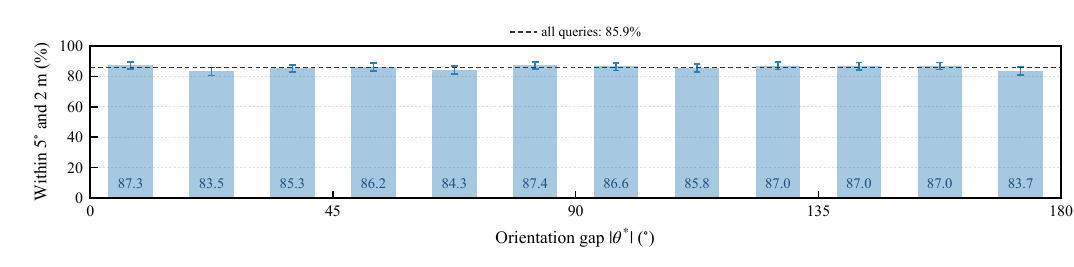}
\end{center}
\caption{
	\textbf{Success rate versus the orientation gap.}
	Fraction of the $9000$ test queries of Bearing-UAV-90K that meet the $5^{\circ}$ and $2$~m tolerance at the returned ($12$th) estimate, in $15^{\circ}$ bins of the ground-truth gap $|\theta^{*}|$, with Wilson $95\%$ intervals; the dashed line marks the overall $85.9\%$.
	All twelve bins are shown rather than the two extremes alone, so that a dip confined to intermediate gaps would still show.
}
\label{fig: orient}
\end{figure*}

%% file: fig/convergence.tex
%

\begin{figure*}[t]
\begin{center}
\includegraphics[width=\widew{1.0}]{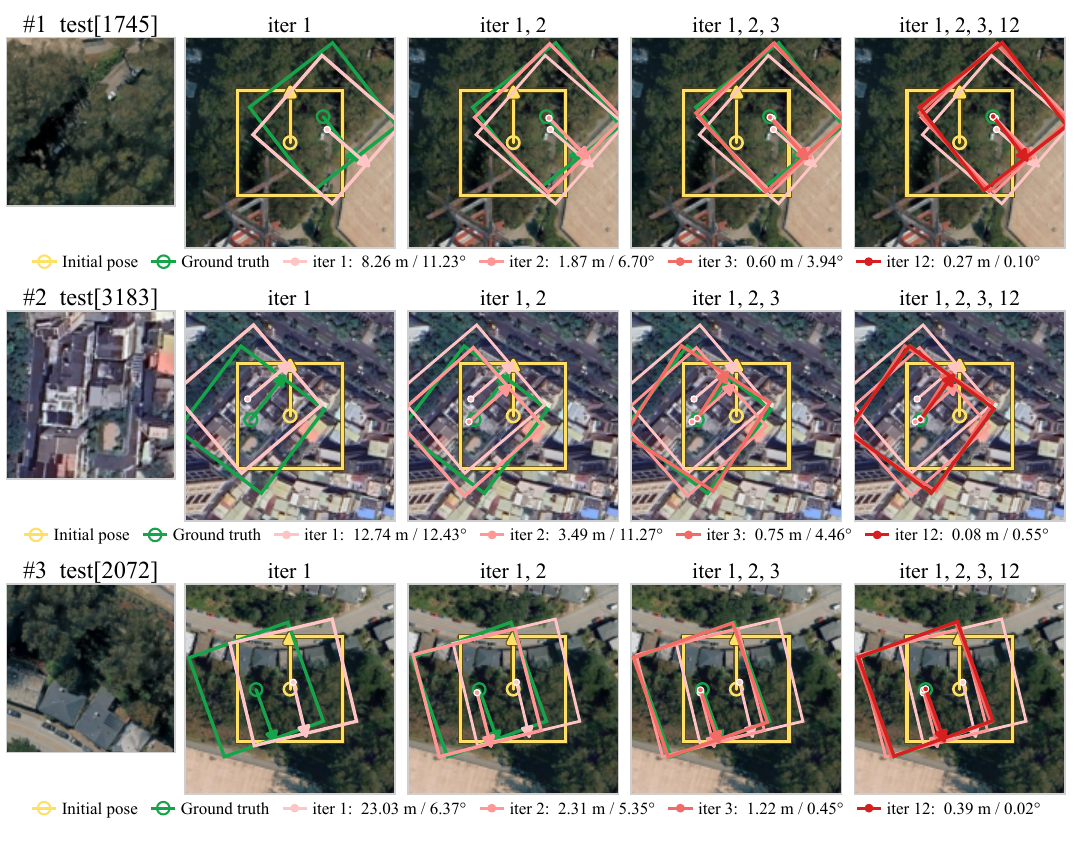}
\end{center}
\caption{
	\textbf{Convergence of the iterative pose estimate.}
	Each row is one test query: the UAV view on the left, followed by the same $512\times512$ satellite block, on which the footprints of the estimates at iterations $1$, $2$, $3$ and $12$ are accumulated over the four panels. The footprint is a square of the same side length as the aerial coverage, drawn at the estimated offset and heading, so its displacement and tilt with respect to the ground truth are the errors of that iteration. Light yellow marks the starting pose (block center, nose pointing north), green the ground truth, and red darkens from iteration $1$ through $2$ and $3$ to $12$. The legend below each row gives the position and heading errors of the four iterations.
}
\label{fig: convergence}
\end{figure*}

%% file: body/tail.tex
\bibliography{wangbo}
%
%